\documentclass[11pt,conference]{ieeeconf}
\IEEEoverridecommandlockouts       
\usepackage{graphics} % for pdf, bitmapped graphics files
\usepackage{epsfig} % for postscript graphics files
\usepackage{mathptmx} % assumes new font selection scheme installed
\usepackage{times} % assumes new font selection scheme installed
\usepackage{amsmath} % assumes amsmath package installed
\usepackage{amssymb}  % assumes amsmath package installed
\usepackage{algorithm} %format of the algorithm
\usepackage{algorithmic} %format of the algorithm
\usepackage{multirow} %multirow for format of table
\usepackage{url}
\usepackage{array}
\usepackage{tabularx}
\usepackage{cite}
\usepackage{booktabs}
\usepackage[table]{xcolor}
\usepackage{pifont}

\newcommand{\cmark}{\ding{51}}
\newcommand{\xmark}{\ding{55}}
\begin{document}

% Change to your title
\title{\LARGE \bf Towards Active Real-to-Twin Inspection: A New Paradigm for Zero-Shot Anomaly Detection}

\author{Jiaxuan Liu$^*$, Yunkang Cao$^*$, Yufeng Chen$^*$, Chunyang Li, Yuhuan Du, and Hui Zhang,~\IEEEmembership{Member,~IEEE} 
\thanks{$^*$Jiaxuan Liu, Yunkang Cao, and Yufeng Chen contributed equally to this work. Jiaxuan Liu, Yunkang Cao, Yufeng Chen, Chunyang Li, Yuhuan Du, and Hui Zhang are with the National Engineering Research Center of Robot Visual Perception and Control Technology, Hunan University, Changsha, Hunan, China (e-mail: ljx317@hnu.edu.cn; zhanghui1983@hnu.edu.cn). Hui Zhang is the corresponding author.}}

\maketitle 
\thispagestyle{empty}

\begin{abstract}

The deployment of zero-shot anomaly detection (AD) in embodied industrial inspection is severely bottlenecked by its reliance on passive, fixed-viewpoint 2D imagery. Such formulations inherently fail to accommodate the active, dynamic observations required in real-world environments. To break this limitation, we introduce Real-to-Twin Anomaly Detection, a novel task that evaluates physical observations directly against geometrically matched CAD Digital Twins. To tackle this new task, we propose AVATAR, \textit{a framework designed to learn robust semantic alignment between Real and Digital Twins}. By bridging benign Sim2Real domain gaps using only defect-free pairs, AVATAR effectively transforms CAD priors into dynamic, anomaly-free references. This elegant formulation enables the model to localize diverse anomalies in a zero-shot manner as unalignable deviations, eliminating the need for defect annotations. Extensive experiments demonstrate that AVATAR substantially outperforms adapted state-of-the-art baselines, exhibiting exceptional robustness to severe viewpoint variations. The code and dataset will be made publicly available.
\end{abstract}

\section{Introduction}

While zero-shot anomaly detection (ZSAD) elegantly identifies defects on unseen objects, its transition to real-world manufacturing exposes notable limitations. Driven by passive 2D visual modeling, current methods often struggle to distinguish true defects from benign environmental noise such as background clutter or illumination changes. More importantly, they overlook the geometric blueprints inherent in Computer-Aided Design (CAD) models. To break this bottleneck, it is highly desirable to advance ZSAD from unconstrained 2D observation to an active perception framework guided by rigorous CAD priors.

\begin{figure}[!t]
\centerline{\includegraphics[width=\columnwidth]{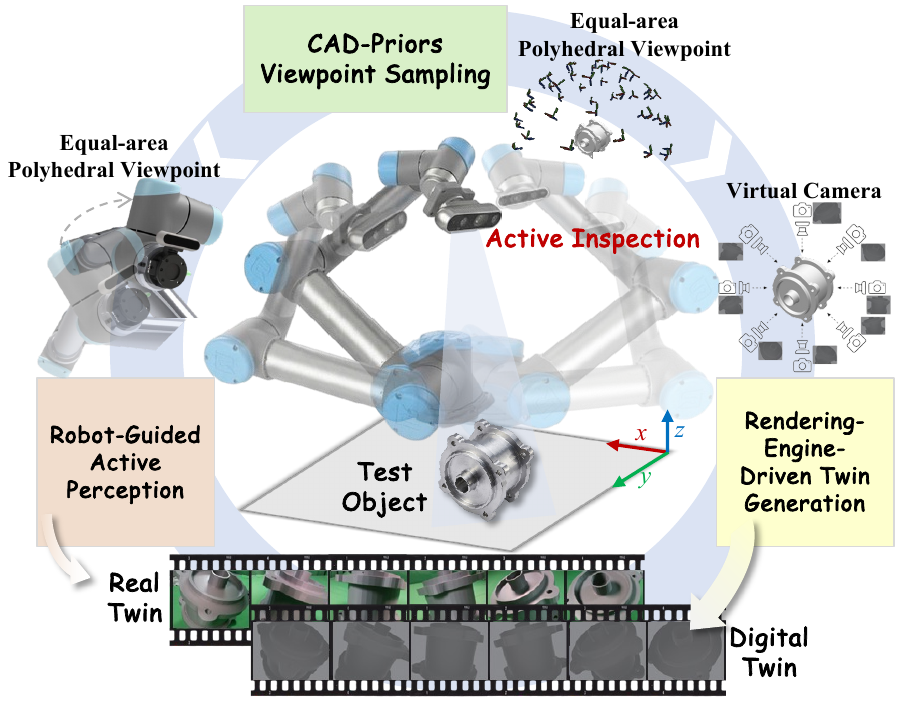}}
    \caption{Overview of the proposed Real-to-Twin inspection pipeline.}
\vspace{-1cm}
\label{fig_1}
\end{figure}

\textit{\textbf{Motivation.}} This perspective highlights two essential ingredients historically absent from ZSAD: a reliable structural blueprint and an active perception mechanism. In modern industrial settings, these elements are naturally embodied by CAD priors and controllable robotic manipulators. This synergy inspires a novel inspection strategy: rather than struggling with decision boundaries over unconstrained 2D images, an embodied agent can actively drive the camera to align with CAD viewpoints. Consequently, anomaly detection transitions into a direct structural verification between physical observations and pose-matched Digital Twins.

\textit{\textbf{Related Work.}} Recent advancements in ZSAD largely rely on adapting powerful vision-language models to industrial contexts \cite{Jeong_2023_CVPR,AdaCLIP,chen2023april,zhou2023anomalyclip,ma2025aa,bayes}. These methods have demonstrated strong performance on static, curated benchmarks such as MVTec \cite{MVTEC} and VisA \cite{VISA}. However, these paradigms are fundamentally confined to passive, 2D image-based inference, rendering them highly fragile to complex viewpoint variations in dynamic, real-world deployments. Concurrently, robotic inspection systems have significantly advanced active viewpoint planning and physical scanning trajectories \cite{robotinspection1,robotinspection2,robotinspection3}. Despite their physical mobility, these systems typically treat perception merely as a supervised downstream task \cite{robotinspection1,robotinspection2}, failing to leverage CAD models as authoritative anchors for design-consistency verification. Recognizing the viewpoint vulnerabilities of 2D models, a parallel line of research has shifted towards multi-view, 3D, and pose-varying paradigms \cite{zhou2023pad, kruse2024splatpose}. Notably, recent benchmarks in this space have explicitly incorporated CAD priors and synthetic data \cite{sim3d, maack2025pcad, pang2023verification}. While acknowledging the value of geometric priors, these works predominantly treat CAD models passively, formulating inspection as offline pose-robust algorithm design or 3D modeling.

To overcome the bottlenecks of existing paradigms, we argue that robust zero-shot inspection cannot rely on passive 2D matching. Instead, it requires coupling CAD geometric priors with robotic mobility to enforce strict viewpoint alignment, as shown in Fig.~\ref{fig_1}. With the camera pose explicitly anchored, the challenge of zero-shot anomaly detection is reduced to a purely visual problem: bridging the Sim2Real domain gap between the physical observations and the rendered twins.

\begin{figure*}[!t]
    \centering
    \includegraphics[width=\textwidth]{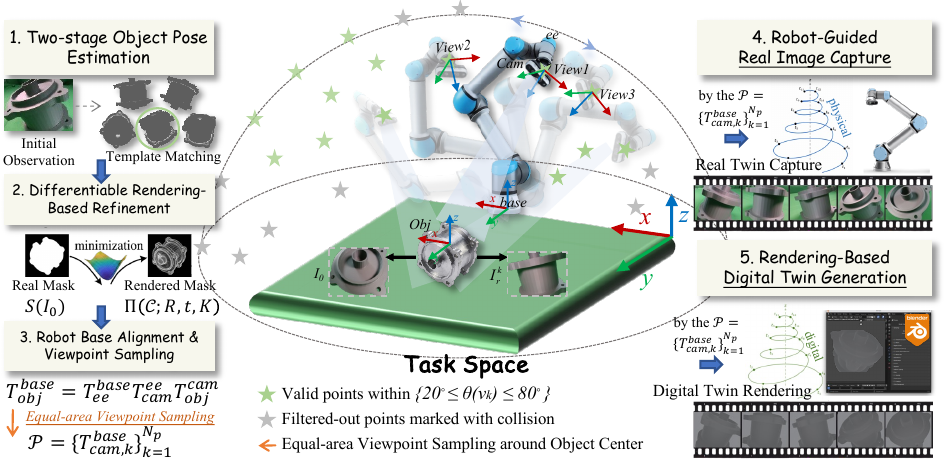}
    \caption{Real-to-Twin data acquisition process for generating spatially aligned real-render image pairs through pose estimation, feasible viewpoint sampling, robotic capture, and synchronized rendering.}
    \vspace{-0.5cm}  
    \label{fig_2}
\end{figure*}

\textit{\textbf{Proposed Framework.}} To address this critical gap, we formalize \textbf{AVATAR} (\textbf{A}ctive \textbf{V}iewpoint \textbf{A}lignment via \textbf{T}win-\textbf{A}nchored \textbf{R}epresentations), a CAD-grounded framework \textit{bridging digital twins and physical reality} for deployable zero-shot anomaly detection. Specifically, driven by CAD geometric priors, a robotic arm executes deterministic viewpoint planning to capture physical observations (the \textit{Real Twin}). Concurrently, a rendering engine synthesizes perfectly pose-matched reference images under the exact same configuration (the \textit{Digital Twin}). Crucially, rather than overfitting to specific object features, AVATAR learns a category-agnostic Sim2Real alignment on available base objects. By mastering this universal mapping from digital renderings to physical reality, the model generalizes seamlessly to entirely unseen targets. This generalized capability transforms any novel Digital Twin into a robust, anomaly-free anchor, enabling highly reliable zero-shot defect localization.

% Our main contributions are threefold:
% \begin{itemize}
%     \item \textbf{Robotic Inspection System:} We construct a complete real-to-twin robotic inspection system. By executing geometry-driven deterministic viewpoint planning, it captures strictly paired real-and-rendered data, physically eliminating the unconstrained pose gap.
%     \item \textbf{Zero-Shot Algorithmic Paradigm:} We propose \textbf{AVATAR}, a novel CAD-grounded algorithmic framework for zero-shot anomaly detection. It employs a twin-anchored alignment adapter to bridge the Sim2Real gap strictly on normal representations, establishing the digital twin as an anomaly-free anchor for precise residual defect localization.
%     % \item \textbf{Dataset \& Open Source:} We introduce the first active Real-to-Twin industrial dataset, and the project resources are publicly available at \url{https://github.com/Jessejx/26_Project_Florence}.
% \end{itemize}

\section{Task Definition and Data Preparation}

% \textbf{Problem Formulation.} We define Real-to-Twin zero-shot anomaly detection, where each physical observation is evaluated against a pose-matched CAD-derived digital reference. Unlike conventional ZSAD that detects deviations from a learned normal distribution, this method defines anomalies as design-inconsistent deviations between the Real Twin and the Digital Twin. The model is trained only on normal real-render pairs from base categories and evaluated on disjoint novel categories without using any test-category images or anomalous samples. 

\textbf{Problem Formulation.} Let $\mathcal{D}_{\mathrm{train}} = \{(I_r^i, I_s^i)\}_{i=1}^{N_{\mathrm{train}}}$ denote the training set, where $(I_r^i, I_s^i)$ represents the $i$-th strictly pose-aligned data pair, consisting of a real physical observation $I_r^i$ and its corresponding CAD-rendered reference $I_s^i$. \textit{This set contains only normal pairs from base object categories to learn category-agnostic cross-domain alignment.} At inference time, the model is evaluated on $\mathcal{D}_{\mathrm{test}} = \{(I_r^j, I_s^j)\}_{j=1}^{N_{\mathrm{test}}}$, comprising unseen novel instances with various anomalies. The training and test categories are strictly disjoint: the model is trained only on normal real-render pairs from base categories, while each test CAD rendering is used only as a design reference at inference. The overall data acquisition process is illustrated in Fig.~\ref{fig_2}.

\textbf{Pose-Grounded Real-Render Pair Construction.} Establishing strict spatial correspondence requires anchoring the physical object's 6D pose within the robotic base frame. Given an initial unconstrained observation $I_0$ and the corresponding end-effector pose, the object pose is estimated via a two-stage procedure. First, a coarse pose candidate $(R_c, t_c)$ is identified through offline template matching between $I_0$ and CAD-rendered views. This estimate is refined using differentiable rendering by minimizing the mask discrepancy:
\begin{equation}
    (R^\ast,t^\ast)=\arg\min_{R,t}\mathcal{L}\big(S(I_0),\Pi(\mathcal{C};R,t,K)\big),
    \label{eq:1}
\end{equation}
\noindent where $S(I_0)$ denotes the foreground mask extracted from the real image $I_0$, and $\Pi(\mathcal{C};R,t,K)$ represents the rendered mask of the CAD model $\mathcal{C}$ under camera intrinsics $K$. These optimized parameters $(R^\ast,t^\ast)$ constitute the object pose within the camera coordinate system, denoted as $T_{\mathrm{obj}}^{\mathrm{cam}}$. Furthermore, by integrating the robot end-effector pose ($T_{\mathrm{ee}}^{\mathrm{base}}$) and the pre-calibrated hand-eye transformation ($T_{\mathrm{cam}}^{\mathrm{ee}}$), the deterministic object pose in the robot base frame is derived as:

\begin{equation}
    T_{\mathrm{obj}}^{\mathrm{base}}=T_{\mathrm{ee}}^{\mathrm{base}}T_{\mathrm{cam}}^{\mathrm{ee}}T_{\mathrm{obj}}^{\mathrm{cam}}.
    \label{eq:2}
\end{equation}
\noindent This alignment serves as the prerequisite for generating physically grounded Real-Digital Twins.

Based on $T_{\mathrm{obj}}^{\mathrm{base}}$, we pre-compute a deterministic trajectory of standard observation viewpoints. We sample an equal-area polyhedral viewpoint set around the object center $c_{\mathrm{obj}}$, filtering out physically infeasible poses (e.g., collisions with the table or workbench) by restricting the elevation angle: $\mathcal{V}=\{v_k \mid 20^\circ \le \theta(v_k)\le 80^\circ\}$. For a given observation radius $r$, this yields a valid camera pose set $\mathcal{P}=\{T_{\mathrm{cam},k}^{\mathrm{base}}\}_{k=1}^{N_p}$. Finally, the robotic arm autonomously executes this planned trajectory. For each pose $T_{\mathrm{cam},k}^{\mathrm{base}}$, the physical camera captures the Real Twin $I_r^k$, while the rendering engine synthesizes the exact Digital Twin $I_s^k$ using the identical spatial configuration $(\mathcal{C}, T_{\mathrm{obj}}^{\mathrm{base}}, T_{\mathrm{cam},k}^{\mathrm{base}}, K)$. Consequently, this rigorous pipeline naturally produces large-scale, one-to-one spatially aligned Real-Digital image pairs $(I_r, I_s)$ that serve as the foundational input for our AVATAR framework.

\begin{figure}[!t]
    \centering
    \includegraphics[width=\columnwidth]{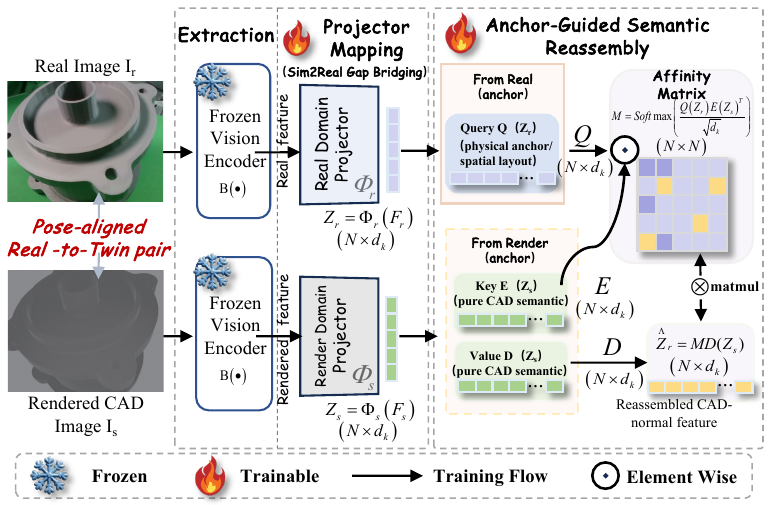}
    \caption{Overview of Twin-Anchored Representation Calibration.}
    \vspace{-0.5cm}  
    \label{fig_3}
\end{figure}

\section{Methodology}
\subsection{Active Real-to-Twin Alignment Overview}
As detailed in Section 2, the robotic-render pipeline provides strictly pose-aligned Real-Digital pairs $(I_r, I_s)$. In principle, pixel-wise subtraction should reveal structural defects; in practice, however, the discrepancy between $I_r$ and $I_s$ also contains (1) \textit{normal domain discrepancy} from illumination, specular reflections, and material variations, and (2) \textit{residual spatial misalignment} caused by robotic jitter and calibration errors. AVATAR therefore learns calibration only from normal pairs to suppress these factors, so that residual responses reflect only \textit{true design inconsistencies}.

\subsection{Twin-Anchored Representation Calibration}

Aligning real observations with CAD-rendered twins is challenging because Real-to-Twin matching must simultaneously handle large Sim2Real appearance gaps (e.g., machining marks, specular reflections, and lighting variations) and residual viewpoint distortions caused by robotic jitter. Under such interference, pixel-wise registration, edge matching, and rigid transformations are unreliable, while direct feature fusion would contaminate the clean digital reference with physical noise. We therefore cast alignment as a dense dictionary retrieval problem, termed \textit{Anchor-Guided Semantic Reassembly}: the real image serves as a spatial anchor, and the digital twin serves as an anomaly-free semantic dictionary. The goal is to retrieve clean digital semantics and reassemble them onto the real geometry.

To instantiate this viewpoint, we first extract dense patch tokens $F_r, F_s$ using a frozen vision foundation model $\mathcal{B}(\cdot)$. To bridge the raw domain gap, dual lightweight projectors non-linearly map these tokens into a shared semantic embedding space, yielding $Z_r = \phi_r(F_r)$ and $Z_s = \phi_s(F_s)$. With the features residing in a shared space, we execute the semantic reassembly. We first compute a dense semantic affinity matrix $\mathcal{M} \in \mathbb{R}^{N \times N}$ between the physical layout and the digital dictionary:
\begin{equation}
    \mathcal{M} = \text{Softmax}\left(\frac{\mathcal{Q}(Z_r) \mathcal{E}(Z_s)^T}{\sqrt{d}}\right),
    \label{eq:3}
\end{equation}
\noindent where $\mathcal{Q}(\cdot)$ extracts spatial topological queries from the real domain, and $\mathcal{E}(\cdot)$ extracts semantic keys from the digital domain. This affinity matrix establishes a differentiable bipartite correspondence, essentially calculating where each real patch should query the digital dictionary to retrieve its corresponding pure feature. Once the correspondences are established, the ideal digital features—encoded by a dictionary projection $\mathcal{D}(Z_s)$—are dynamically routed into the physical coordinate system:
\begin{equation}
    \hat{Z}_s = \mathcal{M} \mathcal{D}(Z_s).
    \label{eq:4}
\end{equation}
\noindent Crucially, this operation bypasses rigid geometric constraints. It dynamically resamples the pure digital reference to adaptively align with the distorted physical observation. As a result, the reconstructed digital twin $\hat{Z}_s$ approximates the layout of the physical observation ($Z_r$), yet is composed entirely of uncontaminated CAD information, providing a rigorous CAD-anchored reference for residual discovery.

\subsection{Optimization and Zero-Shot Scoring}
Trained exclusively on normal Real-to-Twin pairs, AVATAR minimizes the feature discrepancy between the physical anchor and the reassembled digital twin. To prevent gradient contamination from invalid CAD background regions, we employ a foreground-guided local consistency loss:
\begin{equation}
    \mathcal{L}_{\mathrm{local}} = \frac{\sum (1 - \text{CosSim}(Z_r, \hat{Z}_s)) \odot W}{\sum W},
    \label{eq:5}
\end{equation}
\noindent where $\text{CosSim}(\cdot)$ is the channel-wise cosine similarity and $W \in \{0,1\}^{N}$ is the flattened binary rendering mask.

\begin{figure*}[!t]
    \centering
    \includegraphics[width=\textwidth]{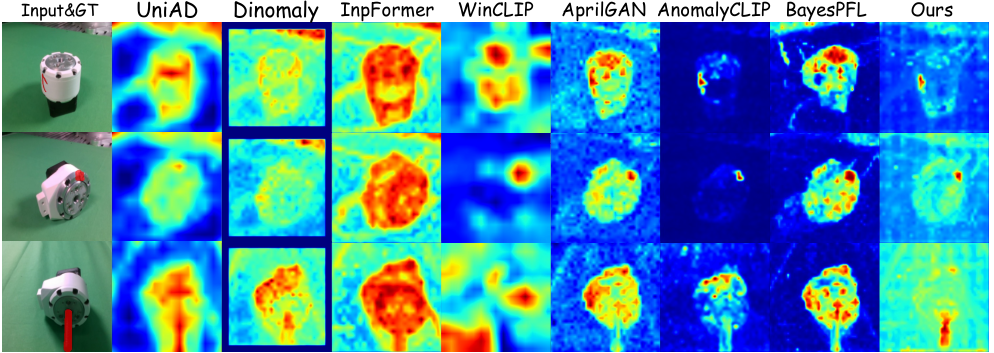}
    \caption{Qualitative anomaly localization results of zero-shot anomaly detection methods.}
    \vspace{-0.2cm}  
    \label{fig_4}
\end{figure*}

However, optimizing $\mathcal{L}_{\mathrm{local}}$ alone risks \textit{representation collapse}: the routing matrix $\mathcal{M}$ may degenerate into an identity mapping, causing $\hat{Z}_s$ to trivially copy $Z_r$ and absorb physical anomalies. To explicitly anchor the reassembled features to their pure CAD semantics, we impose a global regularization loss:
\begin{equation}
\mathcal{L}_{\mathrm{global}} = 1 - \text{CosSim}\big(\text{GAP}(\hat{Z}_s), \text{GAP}(Z_s)\big),
\label{eq:6}
\end{equation}
\noindent where $\mathrm{GAP}(\cdot)$ is global average pooling. The total training objective is $\mathcal{L}_{\mathrm{total}} = \lambda_{local}\mathcal{L}_{\mathrm{local}} + \lambda_{global}\mathcal{L}_{\mathrm{global}}$. During inference, the network reconstructs only normal dictionary features, so structurally anomalous regions yield high residuals. The foreground-masked local discrepancy is therefore used as the zero-shot anomaly score map:
\begin{equation}
\mathcal{A} = (1 - \text{CosSim}(Z_r, \hat{Z}_s)) \odot W.
\label{eq:7}
\end{equation}
\noindent The score map $\mathcal{A} \in \mathbb{R}^{H_F \times W_F}$ is finally upsampled to the original image resolution for precise defect localization.

\section{Experiments}

\subsection{Experimental Settings}
\textbf{Datasets:} We construct a Real-to-Twin (R2T) dataset with strictly paired real-render observations under a category-disjoint zero-shot protocol. The training set contains 216 normal pairs of flange-bearing seats for learning category-agnostic Real-to-Twin alignment, while the test set contains 214 pairs of a robotic end-effector joint system (J6), which serves as the novel category, including normal, texture, structural, and logical anomalies. Pixel-level masks are provided for all anomalous samples. This split prevents object-specific memorization and better reflects practical inspection scenarios.

\textbf{Metrics:} Following prior anomaly detection works~\cite{Jeong_2023_CVPR, AdaCLIP}, we report I-AUROC, I-AP, and I-F1$_{\max}$ for image-level detection, and P-AUROC, P-AP, P-F1$_{\max}$, and AUPRO for pixel-level localization.

\textbf{Implementation Details:} Real images are captured by a UR robotic arm with an Intel RealSense D435 RGB camera, and pose-matched digital references are rendered in Blender under the same camera-object configuration. All real-render pairs are resized to $504 \times 504$. We use DINOv2-ViT-S/14 as a frozen backbone and optimize only the lightweight projection and alignment modules. Unless otherwise specified, the model is trained for 50 epochs with a batch size of 8 using AdamW (learning rate $1\times10^{-4}$, weight decay $1\times10^{-4}$). The local and global alignment losses are equally weighted with $\lambda_{\mathrm{local}}=\lambda_{\mathrm{global}}=1.0$, and the attention module uses a single head.

\subsection{Main Results}

\newcolumntype{C}[1]{>{\centering\arraybackslash}p{#1}}
\newcolumntype{M}[1]{>{\centering\arraybackslash}p{#1}}
\newcolumntype{N}{>{\centering\arraybackslash}X}

\begin{table*}[t]
\centering
\small
\setlength{\tabcolsep}{3.5pt}

\caption{\textbf{Quantitative comparison of zero-shot anomaly detection methods on the proposed Real-to-Twin setting.}}
\label{table_1}

\begin{tabularx}{\textwidth}{@{}C{0.035\textwidth}|M{0.275\textwidth}||NNN|NNNN@{}}
\toprule
& \textbf{\textit{Compared Methods}}
& \multicolumn{3}{c|}{\textbf{\textit{Image-Level}}} 
& \multicolumn{4}{c}{\textbf{\textit{Pixel-Level}}} \\
\cmidrule(lr){1-2} \cmidrule(lr){3-5} \cmidrule(lr){6-9}
\textbf{\scriptsize Type} 
& \textbf{\scriptsize Method} 
& \textbf{\scriptsize I-AUROC} 
& \textbf{\scriptsize I-AP} 
& \textbf{\scriptsize I-F1} 
& \textbf{\scriptsize P-AUROC} 
& \textbf{\scriptsize P-AP} 
& \textbf{\scriptsize P-F1} 
& \textbf{\scriptsize AUPRO} \\
\midrule

\textbf{\textit{T}} 
& Norm. RGB \textit{Residual}         
& 47.88 & 74.66 & \underline{89.01} 
& 79.20 & 1.54 & 4.34 & 51.15 \\

\textbf{\textit{T}} 
& Gradient \textit{Residual}         
& 62.25 & 85.84 & 88.08 
& 85.94 & 3.14 & 7.26 & 78.04 \\

\textbf{\textit{T}} 
& SSIM \textit{Residual}       
& 59.64 & 83.78 & 88.54
& \cellcolor[HTML]{ECF4FF}\textbf{93.93} & 6.70 & 15.64 & 78.34 \\

\textbf{\textit{U}} 
& UniAD\cite{you2022unified} (\small{\textit{NeurIPS-22}})         
& 57.29 & 83.99 & 88.08 
& \underline{93.69} & \underline{15.13} & \underline{22.57} & 65.04 \\

\textbf{\textit{U}} 
& Dinomaly\cite{Dinomaly} (\small{\textit{CVPR-25}})         
& 37.85 & 77.49 & 88.08 
& 77.47 & 2.83 & 5.17 & 67.17 \\

\textbf{\textit{U}} 
& INP-Former\cite{INP} (\small{\textit{CVPR-25}})         
& 31.76 & 71.08 & 88.08 
& 86.33 & 3.17 & 6.53 & 70.58 \\

\textbf{\textit{Z}} 
& WinCLIP\cite{Jeong_2023_CVPR} (\small{\textit{CVPR-23}})         
& 67.31 & 89.23 & 88.08 
& 59.54 & 1.23 & 4.03 & 33.55 \\

\textbf{\textit{Z}} 
& AdaCLIP\cite{AdaCLIP} (\small{\textit{ECCV-24}})        
& 38.26 & 70.52 & 88.08 
& 66.15 & 1.02 & 2.66 & 31.15 \\

\textbf{\textit{Z}} 
& APRIL-GAN\cite{chen2023april} (\small{\textit{CVPR-23}})     
& 61.07 & 84.92 & 88.54 
& 88.33 & 5.62 & 8.84 & \underline{78.85} \\

\textbf{\textit{Z}} 
& AnomalyCLIP\cite{zhou2023anomalyclip} (\small{\textit{ICLR-24}})
& \underline{69.72} & \underline{90.05} & 88.89 
& 90.02 & 6.42 & 8.47 & 64.85 \\

\textbf{\textit{Z}} 
& AA-CLIP\cite{ma2025aa} (\small{\textit{CVPR-25}})         
& 39.08 & 77.07 & 88.08 
& 52.61 & 1.43 & 5.70 & 17.06 \\

\textbf{\textit{Z}} 
& Bayes-PFL\cite{bayes} (\small{\textit{CVPR-25}})        
& 64.45 & 84.91 & 89.47
& 91.51 & 6.18 & 8.20 & \cellcolor[HTML]{ECF4FF}\textbf{83.86} \\

\midrule
\textbf{\textit{Z}} 
& AVATAR (\small{\textit{Ours}})     
& \cellcolor[HTML]{ECF4FF}\textbf{71.15} & \cellcolor[HTML]{ECF4FF}\textbf{90.24} & \cellcolor[HTML]{ECF4FF}\textbf{89.84} 
& 90.55 & \cellcolor[HTML]{ECF4FF}\textbf{23.75} & \cellcolor[HTML]{ECF4FF}\textbf{33.07} & 46.57 \\
\bottomrule
\end{tabularx}
\vspace{-0.5cm}
\end{table*}

Since Real-to-Twin ZSAD is a newly formulated task, there is no established baseline under the same setting. We therefore compare AVATAR with three representative groups of methods: (1) traditional image-processing-based baselines; 
(2) recent unsupervised industrial anomaly detection methods, such as UniAD\cite{you2022unified}, Dinomaly\cite{Dinomaly} and INP-Former\cite{INP}, which learn normal representations from training data; 
and (3) zero-shot anomaly detection methods, including WinCLIP\cite{Jeong_2023_CVPR}, AdaCLIP\cite{AdaCLIP}, APRIL-GAN\cite{chen2023april}, AnomalyCLIP\cite{zhou2023anomalyclip}, AA-CLIP\cite{ma2025aa}, and Bayes-PFL\cite{bayes}, which leverage vision-language priors without task-specific anomaly training.

\subsubsection{Qualitative Comparison}

Fig.~\ref{fig_4} shows qualitative results on \textit{texture, structural, and logical anomalies} from top to bottom. Existing methods often produce either diffuse object-level responses or coarse activations unrelated to the true defect regions. UniAD, Dinomaly, and INP-Former tend to highlight large object areas, suggesting that normal Real-to-Twin appearance gaps are easily confused with anomalies. Zero-shot VLM-based methods, including WinCLIP, APRIL-GAN, AnomalyCLIP, and Bayes-PFL, show some localization ability but remain sensitive to object boundaries, background variations, and semantically salient normal regions. By contrast, our method produces compact and consistent anomaly maps across texture, structural, and logical defects, confirming that CAD-anchored Real-to-Twin alignment better separates true design inconsistencies from normal domain gaps.

\subsubsection{Quantitative Comparison}
Table~\ref{table_1} reports the quantitative comparison on the proposed R2T dataset. AVATAR achieves the best image-level results on I-AUROC, I-AP, and I-F1, showing stronger anomaly discrimination under the category-disjoint zero-shot setting. At the pixel level, AVATAR obtains the highest P-AP and P-F1, outperforming the second-best UniAD by 8.62 and 10.50 points, respectively. This indicates that AVATAR is more effective in producing precise anomaly localization maps. Although SSIM Residual and Bayes-PFL achieve higher P-AUROC and AUPRO, this is partly due to the threshold-sweeping nature of these metrics, which may favor broad anomaly coverage. Their much lower P-AP and P-F1 suggest many false-positive responses and less precise localization. In contrast, AVATAR produces more compact and discriminative anomaly maps, better suppressing normal Real-to-Twin discrepancies.

% Although SSIM Residual and Bayes-PFL achieve higher P-ROC and AUPRO, their much lower P-AP and P-F1 indicate coarser, over-expanded responses rather than precise localization. Taken together with the qualitative results, this shows that AVATAR better suppresses normal Real-to-Twin discrepancies and focuses on design-inconsistent regions.

\subsubsection{Ablation Study}
Table~\ref{table_2} studies the contribution of the main components in AVATAR, including the real-domain projector, the render-domain projector (P-Render), and the foreground-guided local consistency loss (Local Loss). Removing either projector degrades the overall performance, indicating that both real-domain adaptation and render-domain calibration are necessary for reducing the Sim2Real gap. In particular, disabling P-Render causes a clear drop in pixel-level performance, suggesting that a calibrated digital reference is essential for accurate localization. The most significant degradation appears when Local Loss is removed, where the pixel-level performance drops from $48.49$ to $0.60$. %This shows that global feature alignment alone cannot support dense anomaly localization.

\begin{table}[!t]
\centering
\caption{Overall ablation study on the R2T dataset.}
\label{table_2}
\begin{tabular}{ccc|cc}
\toprule
\multicolumn{3}{c|}{Modules} & \multicolumn{2}{c}{Metric} \\
\cmidrule(r){1-3} \cmidrule(l){4-5}
P-Real & P-Render & Local Loss & Image Level & Pixel Level \\
\midrule
\xmark & \cmark & \cmark & $73.09\pm0.02$ & $45.79\pm0.11$\\
\cmark & \xmark & \cmark & $77.32\pm0.01$ & $41.30\pm0.09$\\
\cmark & \cmark & \xmark & $72.68\pm0.02$ & $0.60\pm0.01$\\
\cellcolor[HTML]{ECF4FF}\cmark & \cellcolor[HTML]{ECF4FF}\cmark & \cellcolor[HTML]{ECF4FF}\cmark & \cellcolor[HTML]{ECF4FF}$83.74\pm0.01$ & \cellcolor[HTML]{ECF4FF}$48.49\pm0.07$\\
\bottomrule
\vspace{-0.5cm}
\end{tabular}
\end{table}

\section{Conclusion}
We introduce Real-to-Twin zero-shot anomaly detection, where real observations are inspected against pose-matched CAD-rendered references. To solve this task, we propose \textbf{AVATAR}, which combines active viewpoint alignment with twin-anchored representation calibration to bridge the Sim2Real gap and localize design-inconsistent regions. We construct an R2T dataset under a category-disjoint zero-shot protocol. Experiments show that AVATAR outperforms existing baselines, highlighting the promise of CAD-grounded active perception for practical industrial inspection. This work verifies the feasibility of Real-to-Twin ZSAD, and future work will extend it to broader objects, materials, defects, and robotic scenarios.

\bibliographystyle{IEEEtran}
\bibliography{mylib}

\end{document}